\documentclass[10pt,journal]{IEEEtran}
\usepackage{amsmath,amsfonts,amssymb}
\usepackage{algorithmic}
\usepackage{algorithm}
\usepackage{array}
\usepackage[table]{xcolor}
\usepackage[caption=false,font=footnotesize,labelfont=sf,textfont=sf]{subfig}
\usepackage{textcomp}
\usepackage{stfloats}
\usepackage{url}
\usepackage{verbatim}
\usepackage{graphicx}
\usepackage[nocompress]{cite}
\usepackage{xr}
\graphicspath{{figure/}}
\newcommand{\method}{\textit{ReconDreamer\#}}
\definecolor{RDTableHeader}{RGB}{232,232,232}
\definecolor{RDTableHighlight}{RGB}{218,244,250}
\newcommand{\RDtablestyle}{%
  \small\renewcommand{\arraystretch}{1.16}%
  \setlength{\tabcolsep}{6pt}\setlength{\arrayrulewidth}{0.35pt}}
\newcommand{\RDtabletoprule}{\noalign{\hrule height 0.75pt}}
\newcommand{\RDtablebottomrule}{\noalign{\hrule height 0.75pt}}

\newif\ifshowfloats
\ifdefined\TEXTONLY
  \showfloatsfalse
\else
  \showfloatstrue
\fi
\newcommand{\floatref}[1]{\ifshowfloats\ref{#1}\else\ref{full-#1}\fi}

\begin{document}
\bstctlcite{IEEEexample:BSTcontrol}

\title{\method: Self-Evolving Closed-Loop Synergy of Reconstruction and Generation for Dynamic Driving Scenes}

\author{Songling~Liu$^{*}$,
        Guosheng~Zhao$^{*}$,
        Xiaofeng~Wang,
        Chaojun~Ni,
        Guan~Huang,\\
        Jiwen~Lu,
        Xingang~Wang$^{\dagger}$
        and Zheng~Zhu$^{\dagger}$
\thanks{$^{\dagger}$Xingang Wang and Zheng Zhu are co-corresponding authors.}%
\thanks{$^{*}$Songling Liu and Guosheng Zhao contributed equally to this work and
are co-first authors.}%
\thanks{Songling Liu is with the CAS Engineering Laboratory for Intelligent Industrial
Vision, Institute of Automation, Chinese Academy of Sciences, Beijing 100190, China,
and also with the School of Artificial Intelligence, University of Chinese Academy of
Sciences, Beijing 100049, China (e-mail: liusongling2025@ia.ac.cn).}%
\thanks{Xingang Wang is with the CAS Engineering Laboratory for Intelligent Industrial
Vision, Institute of Automation, Chinese Academy of Sciences, Beijing 100190, China
(e-mail: xingang.wang@ia.ac.cn).}%
\thanks{Guosheng Zhao, Xiaofeng Wang, and Jiwen Lu are with the Department of Automation,
Tsinghua University, Beijing 100084, China.}%
\thanks{Zheng Zhu, Chaojun Ni and Guan Huang are with Giga AI,
Beijing, China
(e-mail: zhengzhu@ieee.org).}%
}

\markboth{IEEE Transactions on Pattern Analysis and Machine Intelligence,~Vol.~XX, No.~XX, XXX~2026}%
{S.~Liu \MakeLowercase{\textit{et al.}}: \method: Self-Evolving DriveRestorer and Adaptive Ground Model}


\maketitle
\begin{abstract}
Dynamic driving scene reconstruction has made remarkable strides in recent years by leveraging generative priors, particularly in the quality of novel trajectory rendering. Nevertheless, a significant gap remains between the generated data and real-world sensor observations, severely hindering further improvements in reconstruction fidelity. In this paper, we present \method{}, a self-evolving framework that synergizes reconstruction and generation for dynamic driving scenes, which substantially bridges the gap between synthesized novel trajectory data and recorded real-world observations. Within this self-evolving closed-loop framework, the generative and reconstruction models mutually reinforce each other through iterative refinement, ultimately yielding a significant enhancement in the 3D representation capability of the reconstruction model. Additionally, we introduce a geometry-prior-based adaptive ground model to further enhance the rendering quality of the reconstruction model. To validate the effectiveness of the proposed method, we conduct comprehensive evaluations on Waymo, nuScenes, PandaSet, and EUVS benchmarks. Specifically on Waymo, at a 6-m lateral shift, \method{} improves NTA-IoU and NTL-IoU by 8.8\% and 3.0\%, respectively, and reduces FID by 15.1\% relative to ReconDreamer++.
\end{abstract}

\section{Introduction}\label{sec:introduction}
\IEEEPARstart{R}{obust} driving scene representation is of paramount importance for autonomous driving simulation systems. With advances in Neural Radiance Fields (NeRF)~\cite{nerf} and 3D Gaussian Splatting (3DGS)~\cite{3dgs,4dgs}, substantial progress has been achieved in reconstructing autonomous driving scenes along recorded trajectories. Surface-aware Gaussian representations further improve geometric fidelity by aligning primitives with reconstructed surfaces or planar structures~\cite{sugar, twodgs, pgsr, rogs}. By incorporating generative priors~\cite{ldm,svd}, recent methods~\cite{recondreamer,drivedreamer4d,streetcrafter} have significantly improved scene representation quality under novel trajectories, emerging as a new paradigm for dynamic driving scene reconstruction. Complementary view-augmentation approaches address extrapolation through inverse warping or view-prior guidance~\cite{flexdrive, evpgs, ggs}.

Central to this paradigm is the generative model's ability to synthesize novel-trajectory data aligned with the recorded views. To train such a generative model, existing methods typically simulate degraded videos along the original trajectories, thereby forming training pairs with the recorded real data. For instance, our prior ReconDreamer series~\cite{recondreamer,recondreamerpp} generates degraded-real data pairs by rendering original trajectories through deliberately under-optimized Gaussian Splatting models. FreeSim~\cite{freesim} simulates degradation through hand-crafted perturbation schemes, such as adding noise perturbations to Gaussian primitives and image blending. Nevertheless, these methods, while capable of generating degraded data, produce degradation patterns that are inconsistent with those encountered during actual novel-trajectory rendering. In practice, the degradation in novel-trajectory rendering primarily stems from artifacts induced by lateral viewpoint deviations in Gaussian Splatting models. This misalignment ultimately limits the 3D representation capacity of the reconstruction model.

\ifshowfloats
\begin{figure*}[t]
    \centering
    \includegraphics[width=0.98\textwidth]{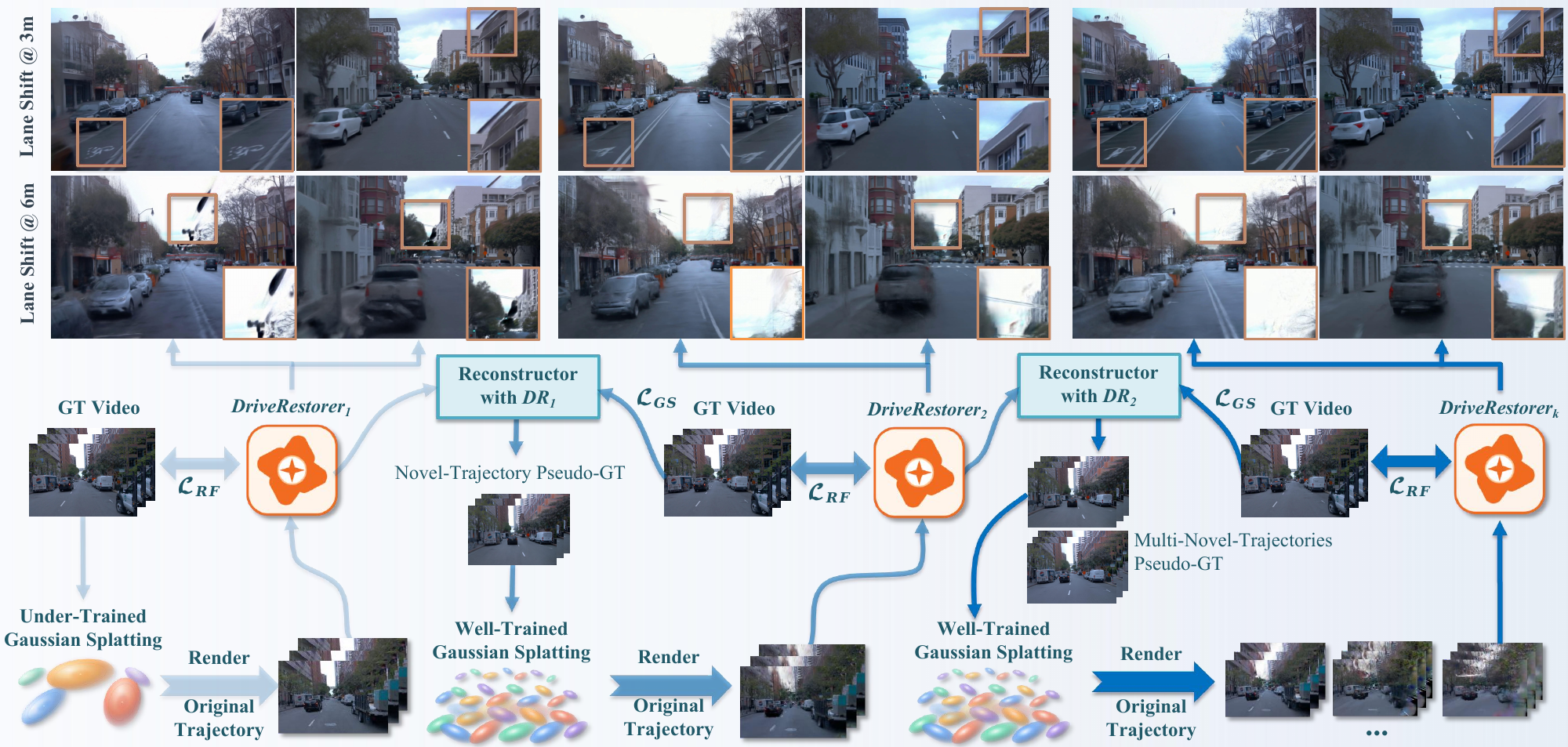}
    \caption{Overview of the self-evolving reconstruction--restoration loop. Starting from recorded video and an under-trained Gaussian scene, DriveRestorer improves novel-trajectory renderings used to train a new reconstruction. Rendering that reconstruction back at recorded poses yields new degraded--real pairs for the next restorer generation. The top rows compare renderings at 3-m and 6-m lane shifts across iterations.}
    \label{fig:teaser}
\end{figure*}
\fi

To alleviate this misalignment, we present \method{}, a self-evolving framework that synergizes reconstruction and generation for dynamic driving scenes. Specifically, as illustrated in Fig.~\ref{fig:teaser}, our framework adopts a multi-round iterative self-evolution architecture. In the first iteration, we follow prior methods~\cite{recondreamer, recondreamerpp} to construct misaligned degraded-real data pairs for training the generative model DriveRestorer, which is then incorporated into the reconstruction pipeline to enhance rendering quality. Subsequently, in the second iteration, the reconstruction model obtained from the first round is employed to render a novel-trajectory video with lateral viewpoint deviations. This video then serves as pseudo ground truth for training a reconstruction model~\cite{streetgs} without DriveRestorer until convergence, after which the trained reconstruction model is applied to render the original trajectory. At this stage, degraded-real data pairs with aligned degradation patterns are obtained. Using such data pairs, we retrain DriveRestorer and subsequently embed it into the reconstruction pipeline. This iterative process is repeated, progressively improving both the generative and reconstruction models in a self-evolving manner. Beyond mitigating the degradation misalignment, we also introduce an Adaptive Ground Model (AGM) to further extend the representation capacity of the ground model in ReconDreamer++~\cite{recondreamerpp}. Instead of relying solely on the hard geometric constraints imposed by LiDAR priors, our approach incorporates regularization-based soft planar constraints on the ground surface. To further validate the effectiveness of the proposed method, we conduct comprehensive evaluations on four widely-used benchmarks: Waymo~\cite{waymo}, nuScenes~\cite{nuscenes}, PandaSet~\cite{pandaset}, and EUVS~\cite{euvs}. Extensive experimental results demonstrate that our method significantly enhances the 3D representation capability of the reconstruction model.

The main contributions of this paper are summarized as follows:
\begin{itemize}
    \item We propose \method{}, a self-evolving framework that synergizes reconstruction and generation for dynamic driving scenes, which substantially mitigates the pattern misalignment issue in generated data pairs.
    \item An Adaptive Ground Model (AGM) is introduced to enhance the representation capacity of the ground model by incorporating soft planar constraints.
    \item Extensive experiments on Waymo, nuScenes, PandaSet, and EUVS demonstrate the superior performance of \method{} along original and novel trajectories. On Waymo, \method{} improves NTA-IoU by up to 8.8\% and NTL-IoU by up to 3.0\%, while reducing FID by up to 15.1\% compared with ReconDreamer++.
\end{itemize}

\noindent\textbf{Extension of the conference versions.}
This manuscript extends our conference papers ReconDreamer~\cite{recondreamer} (CVPR 2025) and ReconDreamer++~\cite{recondreamerpp} (ICCV 2025). It builds on their reconstruction backbone and online DriveRestorer pipeline, together with the scene decomposition, depth supervision, and NTDNet of ReconDreamer++. The journal version introduces two extensions: a self-evolving reconstruction--restoration framework that constructs new degraded--real pairs via auxiliary round-trip reconstruction and updates DriveRestorer across generations, and an Adaptive Ground Model that initializes ground Gaussians from a reconstructed surface and adapts their support, shape, and appearance under local soft planar constraints. We further provide controlled ablations of both extensions, multi-generation DriveRestorer comparisons, and analyses of ground-support coverage and local surface alignment, complementing the evaluation on Waymo, nuScenes, PandaSet, and EUVS. Tab.~\floatref{tab:conference-extensions} summarizes the relationship among these versions.

\ifshowfloats
\begin{table*}[t]\centering\RDtablestyle
\caption{Comparison of the conference versions and the journal extension.}\label{tab:conference-extensions}
\setlength{\tabcolsep}{3pt}
{\small\begin{tabular}{l|ccc|c|ccc|c}\RDtabletoprule
\rowcolor{RDTableHeader}
&\multicolumn{3}{c|}{Base components}&Ground support&\multicolumn{3}{c|}{New mechanisms}&New analyses\\
\rowcolor{RDTableHeader}
Method&\shortstack{Backbone /\\restoration}&\shortstack{Ground\\decomposition}&NTDNet&Policy&\shortstack{Round-trip\\self-evolution}&\shortstack{Surface\\initialization}&\shortstack{Soft planes /\\refinement}&\shortstack{Generation /\\geometry}\\\hline
ReconDreamer~\cite{recondreamer}&$\checkmark$&--&--&\shortstack{Backbone-\\dependent}&--&--&--&--\\
ReconDreamer++~\cite{recondreamerpp}&$\checkmark$&$\checkmark$&$\checkmark$&Fixed&--&--&--&--\\\hline
\rowcolor{RDTableHighlight}
\textbf{\method}&$\checkmark$&$\checkmark$&$\checkmark$&\textbf{Adaptive}&$\checkmark$&$\checkmark$&$\checkmark$&$\checkmark$\\\RDtablebottomrule
\end{tabular}}
\end{table*}
\fi

\section{Related Work}\label{sec:related-work}
\subsection{Neural Reconstruction of Driving Scenes}
Neural radiance fields represent scene appearance through a continuous volumetric function and enable novel-view synthesis from calibrated images~\cite{nerf}. Subsequent methods improve the treatment of large scenes, sampling, and computation. S-NeRF and UC-NeRF address street views and camera calibration~\cite{snerf,ucnerf}, while StreetSurf and deformable neural meshes investigate outdoor geometry~\cite{streetsurf,neuralmesh}. Mip-NeRF 360 handles unbounded environments, while Instant-NGP uses multiresolution feature grids to accelerate representation and optimization~\cite{mipnerf360,instantngp}. Zip-NeRF combines grid-based representations with anti-aliasing, and NeO 360 investigates sparse-view synthesis of outdoor scenes~\cite{zipnerf,neo360}. For driving scenes, Urban Radiance Fields incorporates sensor information into outdoor reconstruction and Block-NeRF partitions large environments into tractable components~\cite{urbanrf,blocknerf}. These approaches demonstrate the importance of representation design and sensor alignment, but observations collected along one path still constrain some viewing directions much more strongly than others.

Dynamic environments require additional structure. Neural Scene Flow Fields and HyperNeRF model temporal changes through scene flow or higher-dimensional deformation, while K-Planes and NeRFPlayer explore factorized and decomposed dynamic radiance representations~\cite{nsff,hypernerf,kplanes,nerfplayer}. ENeRF and HyperReel address efficient free-viewpoint video synthesis through geometry-guided or ray-conditioned sampling~\cite{enerf,hyperreel}. Neural Scene Graphs and Panoptic Neural Fields explicitly associate scene content with objects, while MARS and UniSim study compositional scene modeling and neural sensor simulation in driving settings~\cite{nsg,panopticnf,mars,unisim}. HUGSIM builds a real-time closed-loop driving simulator from reconstructed Gaussian scenes, supporting ego and actor updates driven by control commands~\cite{hugsim}. NeuRAD further develops neural rendering for autonomous-driving sensor configurations~\cite{neurad}. Other approaches incorporate sensor cues or decompose dynamic urban scenes~\cite{lidarf,suds,emernerf}. ADGaussian predicts generalizable Gaussians from image and depth features, while DGGT reconstructs dynamic scenes and camera poses from unposed images~\cite{adgaussian,dggt}. Such systems must model both appearance and the transformations that place agents in the world at each timestamp. A novel trajectory changes how these components occlude one another, even when their underlying motion is unchanged. Reliable simulation therefore requires coherent object placement and background reconstruction rather than independent frame-level plausibility.

3D Gaussian splatting offers an explicit, differentiable representation with efficient rasterization~\cite{3dgs}. Mip-Splatting improves its behavior across changes in sampling scale through anti-aliasing~\cite{mipsplatting}. Dynamic extensions model temporal changes with deformable or 4D Gaussians~\cite{deformablegs,4dgs}. GaussianPro guides densification through progressive geometric propagation, using patch matching to create Gaussians with accurate positions and orientations~\cite{gaussianpro}. Other methods organize Gaussians for adaptive rendering and large scenes~\cite{scaffoldgs,vastgaussian,mvgs}. Street Gaussians separates static content and dynamic objects, enabling urban-scene rendering through object-aware Gaussian representations~\cite{streetgs}. Periodic Vibration Gaussian, S3Gaussian, and OmniRe explore additional ways to represent dynamic urban content and scene decomposition~\cite{pvg,s3gaussian,omnire,dynamic3dgf}. DrivingGaussian, AutoSplat, HO-Gaussian, and HUGS also study Gaussian representations for urban driving scenes~\cite{drivinggaussian,autosplat,hog,hugs}. DrivingRecon investigates a feed-forward reconstruction model for autonomous driving~\cite{drivingrecon}. Their explicit primitives provide convenient interfaces for semantic grouping, geometric constraints, and support refinement. Our method uses this interface to assign a dedicated optimization policy to ground. It does not replace the Gaussian renderer or introduce an alternative dynamic-object model. Instead, it modifies the supervision loop and the way reliable road support is established and maintained.

\subsection{Generative Priors and Trajectory Extrapolation}
Latent diffusion models provide learned appearance priors for image synthesis~\cite{ldm,sdxl}. Video diffusion extends these priors to temporal generation~\cite{svd,emuvideo,imagenvideo,photorealvideo}, including high-motion synthesis~\cite{makepixelsdance}. Video Diffusion Models and Align Your Latents develop temporally coherent video generation with diffusion~\cite{videodiffusion,alignlatents}. Latte and CogVideoX explore transformer-based video diffusion~\cite{latte,cogvideox}, while CogVideo, VideoGPT, and VideoPoet use transformer-based generative models~\cite{cogvideo,videogpt,videopoet}. Driving world models, including the DriveDreamer series, adapt generation to the structure of road environments~\cite{drivedreamer,drivedreamer2,unidrivedreamer}. GAIA-1, Vista, and Drive-WM also investigate driving video generation and world modeling~\cite{gaia1,vista,drivewm}. WorldDreamer and Pandora study broader video-based world models~\cite{worlddreamer,pandora}. DriveArena and MagicDrive3D extend driving generation toward simulation and controllable 3D scene synthesis~\cite{drivearena,magicdrive3d}. MagicDrive, Panacea, and DriveScape explore controllable driving video generation~\cite{magicdrive,panacea,drivescape}. GenAD and DrivingDiffusion study predictive or layout-guided generation for driving scenarios~\cite{genad,drivingdiffusion}. The resulting videos can be used directly as visual simulations or as additional observations for a reconstructive model. These roles have different requirements: an attractive video is not necessarily consistent with the geometry, agent poses, or road layout of a particular reconstructed scene.

DriveDreamer4D investigates generated observations as data for driving-scene representation, and FreeVS studies generative view synthesis along free driving trajectories~\cite{drivedreamer4d,freevs}. DreamDrive generates visual references with video diffusion and lifts them into a hybrid Gaussian representation for 3D-consistent dynamic driving-scene rendering~\cite{dreamdrive}. StreetCrafter and methods using diffusion priors with Gaussian representations further illustrate the interaction between conditional generation and view extrapolation~\cite{streetcrafter,sgd,vegs}. DriveX and RGE-GS further explore generative guidance for extrapolated driving views~\cite{drivex,rgegs}. DIFIX3D+ repairs novel-view renderings with a single-step diffusion model and distills the corrected views back into the scene representation~\cite{difix3dplus}. Generative supervision can fill appearance gaps, but errors can become persistent when they are repeatedly fitted by the scene model. The quality of supervision must consequently be assessed together with the representation into which it is incorporated.

Recent work explores complementary conditioning and fusion strategies. LidarPainter uses one-step diffusion with sparse LiDAR and corrupted renderings~\cite{lidarpainter}. DriveFix models cross-camera and temporal consistency, while SPVC combines structured conditions with panoptic video repair~\cite{drivefix,spvc}. SymDrive uses symmetric views and autoregressive restoration, and DriveExplorer progressively restores extrapolated renderings~\cite{symdrive,driveexplorer}. GenRe instead learns a generalizable enhancer that distills diffusion priors into pretrained Gaussian representations~\cite{dge}.

FaithFusion uses pixel-wise information gain to guide correction and distillation, while PRG-Fusion routes regions among preservation, repair, and generation~\cite{faithfusion,prgfusion}. StreetNVS jointly conditions video diffusion on LiDAR, reference images, and camera poses, and GeoFlow uses geometry-aligned priors for efficient driving video generation~\cite{multisensor,geoflow}. VDEGaussian combines video diffusion with timestamp refinement and uncertainty-aware distillation for dynamic urban scenes~\cite{vdegaussian}. A complementary application is ReconDreamer-RL, which incorporates diffusion-enhanced reconstruction into closed-loop simulation and reinforcement learning~\cite{recondreamerrl}.

ReconDreamer introduces DriveRestorer, a restoration model trained from imperfect renders and recorded video targets, and uses it for online reconstruction improvement~\cite{recondreamer}. ReconDreamer++ addresses the gap between generated and recorded observations through scene decomposition, novel-view depth supervision, and NTDNet~\cite{recondreamerpp}. Our extension focuses on the distribution used to train the restorer itself. Instead of treating that distribution as fixed, the current scene and restorer construct the inputs of the next training round. Real recorded images remain the restoration targets. The auxiliary reconstruction provides a way to exploit novel-view failures without treating generated videos as verified labels for those failures.

\subsection{Surface Reconstruction and Ground Constraints}
Road geometry benefits from priors that are stronger than those appropriate for arbitrary background objects. LiDAR-based ground segmentation identifies measurements that can support such priors~\cite{groundseg}. Surface reconstruction then supplies connectivity and a more regular spatial discretization. We use Field-Aligned Online Surface Reconstruction as the initialization operator~\cite{osr}, followed by Butterfly-style subdivision to refine the mesh sampling~\cite{butterfly}. These established geometric tools are components of our initialization, rather than new surface-reconstruction algorithms proposed by this work. RoGs places 2D Gaussian surfels on a regular meshgrid and initializes their heights and orientations from vehicle poses~\cite{rogs}. ExtraGS combines road-surface Gaussians with an SDF and uncertainty-guided generative supervision, while ViSE uses a dimension-reduced road SDF within a vision-only extrapolation pipeline~\cite{extrags,vise}.

The distinction between surface construction and radiance representation is central. A reconstructed mesh can guide Gaussian locations without becoming the final rendering primitive. After initialization, Gaussian scales, orientations, opacity, and spherical-harmonic coefficients still determine the rendered image. Local plane constraints therefore operate at several levels: they restrict support movement, encourage thin covariance along the surface normal, and discourage novel-trajectory deformations from moving road content off the measured surface. A single position constraint does not by itself control all of these effects.

ReconDreamer++ already separates ground and fixes the base ground positions and primitive count~\cite{recondreamerpp}. AGM extends this treatment with explicit support adaptation and pruning protection. Its local planes accommodate road slope and changing orientation without assuming that the entire road lies on one horizontal plane. At the same time, the formulation retains the limits of the sensor data: incorrect segmentation, curb transitions, or missing measurements can still produce poor local neighborhoods. Our contribution concerns the coordinated use of reliable support, local geometry, and adaptive Gaussian refinement under generative supervision.

\section{Method}\label{sec:method}
\subsection{Preliminary}
ReconDreamer~\cite{recondreamer} proposes the incremental integration of world model knowledge, significantly enhancing the model's performance for large maneuvers. The training process of ReconDreamer is divided into two stages. In the first stage, a partially trained reconstruction model~\cite{streetgs} is utilized to render videos along the original trajectory. This process naturally introduces ghosting artifacts or blurriness, a direct consequence of the model's underfitting. These degraded outputs are then paired with recorded sensor data to form training pairs, which are used to optimize the video restoration model, referred to as DriveRestorer~\cite{recondreamer}. In the second stage, degraded videos of novel trajectories are first rendered, and the fine-tuned DriveRestorer is subsequently employed to refine these rendering results, enhancing both their quality and spatial consistency. These restored outputs, combined with the original video data, are then integrated into the training pipeline to further enhance the performance and robustness of the reconstruction model. This two-stage approach ensures that ReconDreamer achieves superior performance across the novel trajectories. Additionally, ReconDreamer++ adheres to the two-stage paradigm while further optimizing the representation of dynamic driving scenes, thereby improving reconstruction quality.

\subsection{Overview of \method{}}
As illustrated in Fig.~\ref{fig:self-evolving}, \method{} introduces a self-evolving architecture based on multi-stage iterative optimization, which substantially reduces the mismatch between the degraded inputs seen during training and those encountered at inference time. Furthermore, we introduce an Adaptive Ground Model (AGM) to enhance the representational capacity for dynamic driving scenes. These two components are detailed in the following sections.

\ifshowfloats
\begin{figure*}[t]
    \centering
    \includegraphics[width=0.98\textwidth]{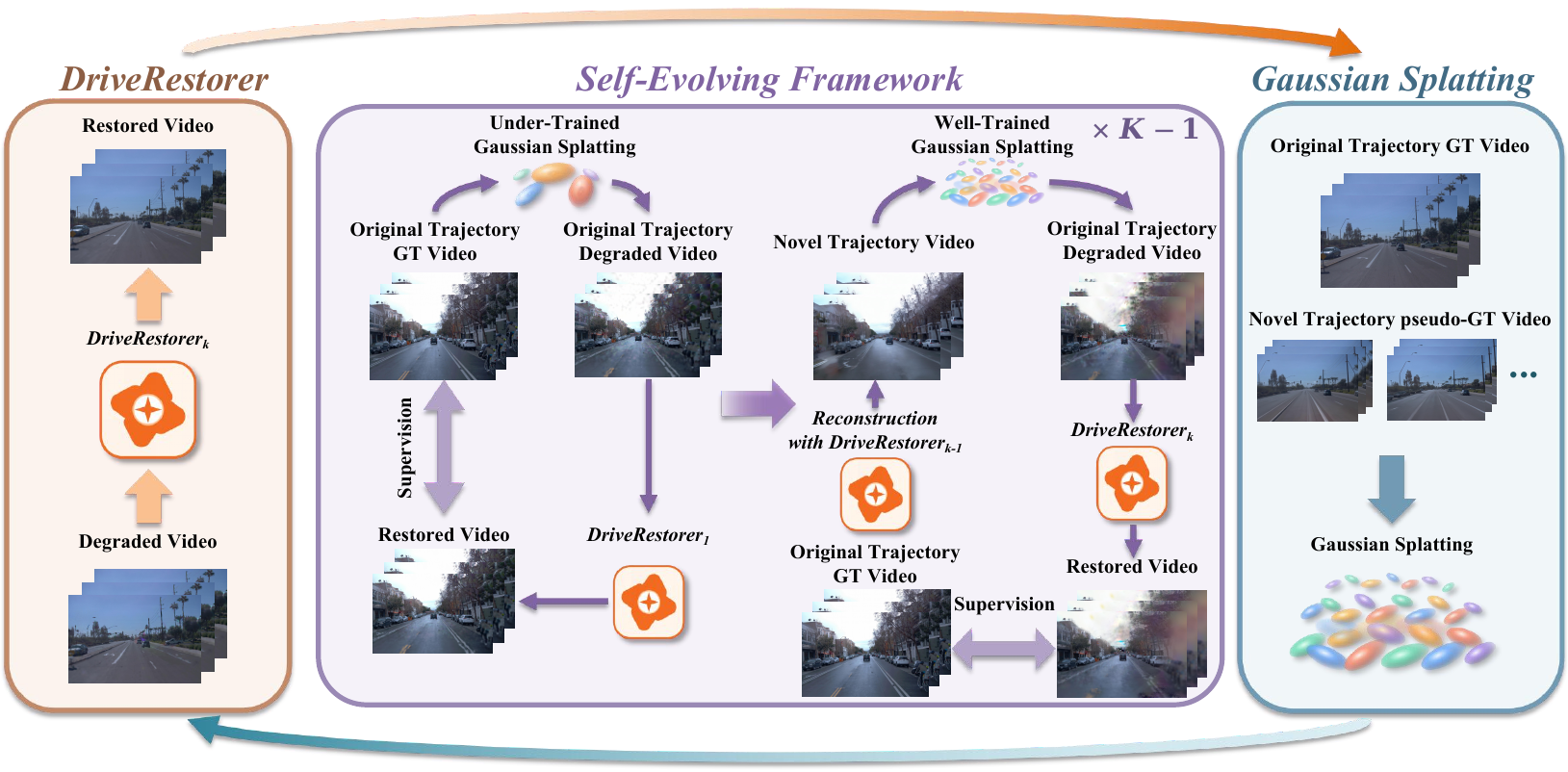}
    \caption{Trajectory round-trip construction for Self-Evolving DriveRestorer. Novel-trajectory rendering and restoration feed an auxiliary reconstruction, which returns to recorded poses to construct inputs paired with real video targets.}
    \label{fig:self-evolving}
\end{figure*}
\fi

\subsection{Self-Evolving Framework}
The proposed self-evolving framework consists of multiple rounds of iteration. Each round comprises three stages: pair construction, restoration model training, and reconstruction model updating. By alternating between updating the restoration model and the reconstruction model, the framework progressively aligns the degradation distribution faced by the restorer with the degradation patterns encountered in practical deployment, thereby expanding its failure coverage over successive rounds. 

\noindent{\textbf{The first round.}} Following ReconDreamer~\cite{recondreamer} and ReconDreamer++~\cite{recondreamerpp}, we construct degraded training pairs from an under-converged Gaussian Splatting model $\mathcal{G'}_{1}$~\cite{streetgs}. The resulting renders along the original trajectories contain reconstruction artifacts and thus serve as degraded inputs for restoration training. These videos, together with the corresponding real observations, form the initial round-trip pair collection $\mathcal{D}_1$, which is then used to train the restoration model DriveRestorer $\mathcal{R}_{1}$. After training, DriveRestorer is incorporated into the ReconDreamer pipeline to provide restored observations for novel trajectories and to participate in the optimization of the main reconstruction model $\mathcal{G}_{1}$. 

\noindent{\textbf{The second round.}} The reconstruction model $\mathcal{G}_{1}$ trained in the first round is used to render novel-trajectory videos, which serve as supervision to train a Gaussian Splatting model $\mathcal{G'}_{2}$ without a generative model. In contrast to the first stage, where the Gaussian Splatting model is deliberately left under-converged, this model $\mathcal{G'}_{2}$ is trained to convergence. Once converged, it is used to render original trajectory videos. Since this Gaussian Splatting model is trained on novel trajectories, rendering the original trajectories exposes it to a viewpoint shift, thereby producing rendering degradations that more closely resemble those encountered in practical deployment. The resulting low-quality rendered videos, together with the original captured data, constitute the second-round pair set $\mathcal{D}_2$. Together with the first-round pair set $\mathcal{D}_1$, they form the training set $\{\mathcal{D}_1, \mathcal{D}_2\}$ for the video restoration model $\mathcal{R}_{2}$ at this stage, which is then incorporated into the reconstruction pipeline to optimize the reconstruction model $\mathcal{G}_{2}$ of this round.

\noindent{\textbf{More rounds.}} On the basis of the above procedure, the framework can be extended to multiple rounds. In each round $k$, the reconstruction model $\mathcal{G}_{k-1}$ from the previous round is used to render novel trajectory videos, which can be regarded as pseudo ground truth for the novel trajectories. These renderings are then used to train an auxiliary Gaussian splatting model $\mathcal{G'}_{k}$.
Since $\mathcal{G'}_{k}$ is trained on novel trajectories, rendering along the original trajectories exposes it to a viewpoint shift, thereby producing trajectory-shifted observations that constitute the pair set $\mathcal{D}_k$ for round $k$. The accumulated pairs $\cup_{j=1}^{k}\mathcal{D}_j$ are then used to train the restoration model $\mathcal{R}_{k}$ for the current round. Once $\mathcal{R}_{k}$ is trained, it is incorporated into the reconstruction pipeline to optimize the reconstruction model $\mathcal{G}_{k}$ of this round. This process keeps the degradation distribution modeled by the restorer synchronized with the current optimization state of the reconstruction model, thereby achieving continuous improvement in restoration performance and reconstruction quality over successive alternating updates. Furthermore, as the iteration progresses, we progressively increase the number of novel trajectories rendered by $\mathcal{G}_{k-1}$, thereby expanding the training data available for $\mathcal{R}_{k}$ and fostering mutual reinforcement between the generative and reconstruction models within the self-evolving framework.

\subsection{Adaptive Ground Model}\label{sec:ground}
To further enhance the 3D representation capability of the dynamic driving-scene reconstruction model, we introduce an Adaptive Ground Model (AGM). Unlike the ground model introduced in ReconDreamer++~\cite{recondreamerpp}, which relies strictly on the acquired ground point cloud and keeps the number and location parameters of ground Gaussians fixed during subsequent optimization, AGM imposes geometric constraints on the ground surface and optimizes all parameters of the ground model during training. 

The overall procedure of AGM is summarized in Algo.~\floatref{alg:agm}. Specifically, let $P_g$ be the segmented ground LiDAR points. We first reconstruct a surface from $P_g$ and then downsample it to the target size~\cite{osr, butterfly}. The resulting vertices of ground become Gaussian means to initialize the ground model $\mathcal G_g$. Each vertex receives appearance, opacity, scale, and rotation parameters through inverse-distance interpolation from nearby members of $\mathcal G_g$; quaternion results are normalized.

After initialization, the ground component is optimized under a geometric regularizer. Writing $r_{ij}$ for a rotation axis, $q_i=\arg\min_j s_{ij}$ for the thinnest axis, and $b_i=\arg\max_j|n_i^\top r_{ij}|$ for the most normal-aligned axis, the ground regularizer is
\begin{equation}
\begin{aligned}
\mathcal L_g={}&\frac{\lambda_p}{N_g}\sum_i
 |n_i^\top(\widehat\mu_i(C)-p_i)|\\
&+\frac{\lambda_o}{N_g}\sum_i(1-|n_i^\top r_{i q_i}|)
+\frac{\lambda_f}{N_g}\sum_i s_{i b_i}.
\end{aligned}\label{eq:groundloss}
\end{equation}
Here $\widehat\mu_i(C)$ is the rendered center under the enabled novel-view deformation. A geometric projection maintains base support, whereas $\mathcal L_g$ constrains rendered centers and Gaussian shape; the two operations are distinct.

For support refinement, AGM retains the backbone's valid image-gradient and size criteria and applies different operations to ground primitives depending on their scale: large primitives are split, whereas small ones are duplicated. For large primitives, a selected parent is split by placing a child at
\begin{equation}
\mu_{i,l}^{\rm child}=\mu_i+(I-n_i n_i^\top)u_{i,l}.
\label{eq:tangentsplit}
\end{equation}
where $u_{i,l}$ is an offset sampled from the parent ellipsoid. The projection removes the normal component of this offset, so that the child remains on the same local plane as its parent. Split children inherit the backbone's attributes and scale shrinkage. For small primitives, AGM applies duplication, following the backbone's ordinary duplication rule without additional geometric modification. Both operations alter the locations of ground primitives, so the updated means are no longer guaranteed to lie on the measured surface. We therefore introduce an additional plane-maintenance operation. For each current ground center $\mu_i$, the centroid $p_i$ and least-variance normal $n_i$ are estimated from a local neighborhood of $P_g$~\cite{groundseg}. With $\|n_i\|=1$, an enabled plane-maintenance operation projects the base mean by
\begin{equation}
\mu_i\leftarrow\mu_i-n_i n_i^\top(\mu_i-p_i).
\label{eq:projection}
\end{equation}

\ifshowfloats
\begin{algorithm}[t]
\caption{Adaptive Ground Model}
\label{alg:agm}
\footnotesize
\begin{algorithmic}[1]
\REQUIRE Ground LiDAR $P_g$; initial Gaussians $\mathcal G_g^0$; scene-training observations $\mathcal O$
\ENSURE Ground representation $\mathcal G_g$
\STATE Reconstruct a surface from $P_g$ and downsample it to the target size
\STATE Initialize ground means from the resulting surface vertices
\STATE Interpolate appearance, opacity, scale, and rotation from $\mathcal G_g^0$ by inverse distance; normalize quaternions
\STATE Initialize the optimizer over all ground parameters, including means
\FOR{each scene-optimization iteration $t$}
  \STATE Sample an observation with camera $C$ from $\mathcal O$
  \STATE Estimate local centroids $p_i$ and unit normals $n_i$ from $P_g$ at current means
  \STATE Compute rendered centers $\widehat\mu_i(C)$ and axis indices $q_i,b_i$; evaluate $\mathcal L_g$ by Eq.~\eqref{eq:groundloss}
  \STATE Optimize all ground parameters with $\mathcal L_{\rm rec}+\mathcal L_g$
  \STATE Accumulate valid image-gradient and screen-size statistics
  \IF{a backbone support-refinement step is due}
    \STATE Select large primitives to split and small primitives to duplicate using the backbone's gradient and size criteria
    \STATE Split large primitives by Eq.~\eqref{eq:tangentsplit}; inherit attributes and scale shrinkage
    \STATE Duplicate small primitives using the backbone's ordinary rule
    \STATE Synchronize ground attributes, optimizer state, and refinement statistics
  \ENDIF
  \IF{plane maintenance is enabled at iteration $t$}
    \STATE Re-estimate $(p_i,n_i)$ from $P_g$ at current ground means
    \STATE Project ground means onto the local planes by Eq.~\eqref{eq:projection}
  \ENDIF
\ENDFOR
\end{algorithmic}
\end{algorithm}
\fi

\section{Experiments}\label{sec:experiments}
\ifshowfloats
\begin{table*}[t]\centering\RDtablestyle
\caption{Quantitative comparison on Waymo. NTA and NTL denote NTA-IoU and NTL-IoU, respectively.}\label{tab:waymo}
{\small\begin{tabular}{l|rrr|rrr|rrr}\RDtabletoprule
\rowcolor{RDTableHeader}
&\multicolumn{3}{c|}{Recorded trajectory}&\multicolumn{3}{c|}{Lateral shift: 3 m}&\multicolumn{3}{c}{Lateral shift: 6 m}\\
\rowcolor{RDTableHeader}
Method&PSNR$\uparrow$&SSIM$\uparrow$&LPIPS$\downarrow$&NTA$\uparrow$&NTL$\uparrow$&FID$\downarrow$&NTA$\uparrow$&NTL$\uparrow$&FID$\downarrow$\\\hline
Street Gaussians~\cite{streetgs}&36.50&0.9570&0.1149&0.4980&53.19&130.75&0.3740&49.27&213.04\\
FreeVS~\cite{freevs}&26.59&0.8057&0.1325&0.5050&56.84&104.23&0.4650&55.37&121.44\\
DriveDreamer4D~\cite{drivedreamer4d}&34.37&0.9447&0.1321&0.4570&53.30&113.45&0.1590&50.09&261.81\\
ReconDreamer~\cite{recondreamer}&34.31&0.9431&0.1519&0.5390&54.58&93.56&0.4670&52.58&149.19\\
ReconDreamer++~\cite{recondreamerpp}&36.29&0.9573&0.1076&0.5720&57.06&72.02&0.4890&56.57&111.92\\\hline
\rowcolor{RDTableHighlight}
\textbf{\method}&\textbf{36.89}&\textbf{0.9638}&\textbf{0.0953}&\textbf{0.5984}&\textbf{58.75}&\textbf{66.42}&\textbf{0.5320}&\textbf{58.29}&\textbf{94.99}\\\RDtablebottomrule
\end{tabular}}
\end{table*}
\begin{figure*}[t]
    \centering
    \includegraphics[width=0.98\textwidth]{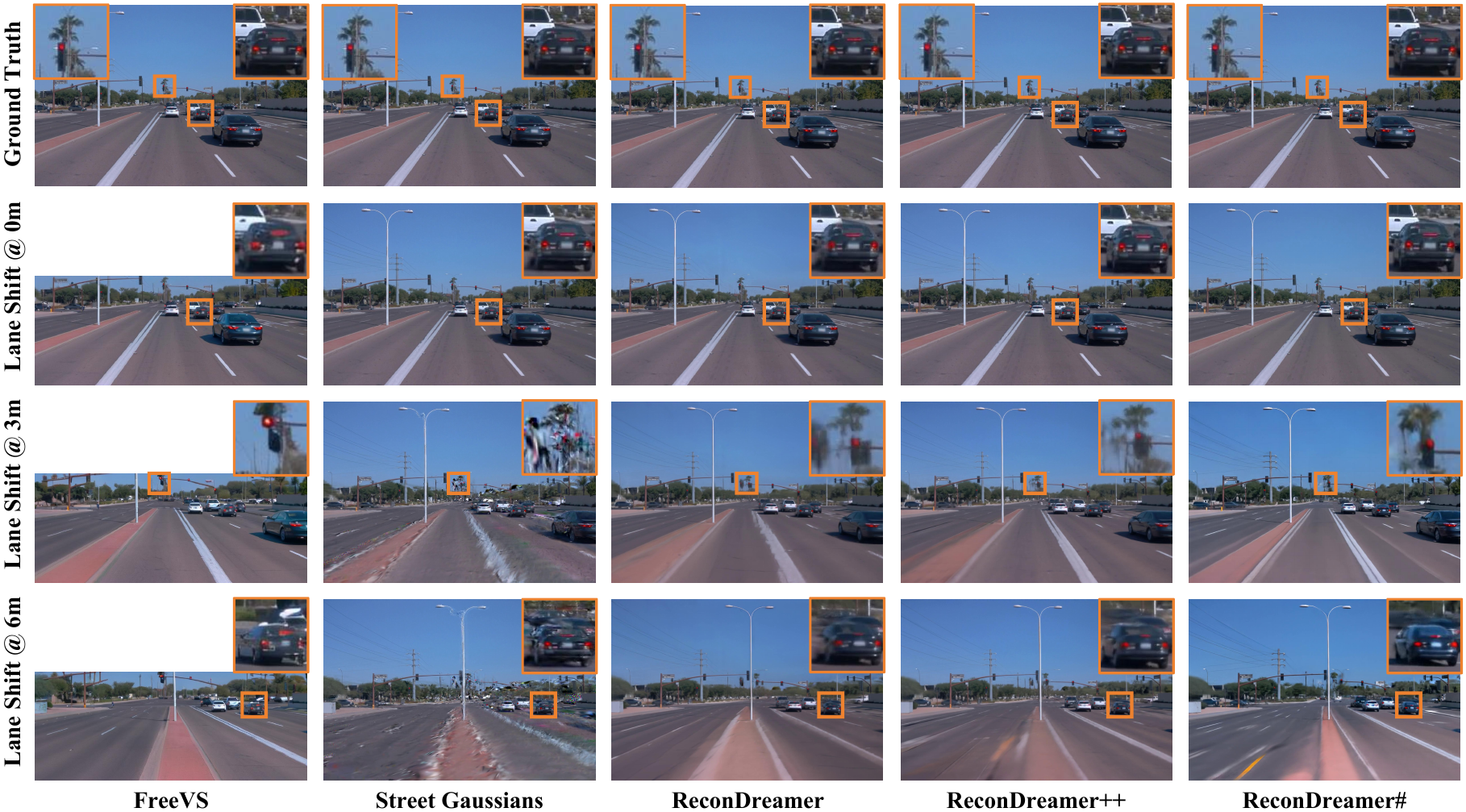}
    \caption{Qualitative comparison on Waymo. Columns show FreeVS, Street Gaussians, ReconDreamer, ReconDreamer++, and \method{}; rows show the recorded-view reference and renderings at 0-m, 3-m, and 6-m lateral shifts. Orange boxes enlarge vehicle and roadside details. Ground truth is shown only for the recorded view.}
    \label{fig:waymo}
\end{figure*}
\begin{table*}[t]\centering\RDtablestyle
\caption{Quantitative comparison on nuScenes.}\label{tab:nuscenes}
{\small\begin{tabular}{l|rrr|rrr|rrr}\RDtabletoprule
\rowcolor{RDTableHeader}
&\multicolumn{3}{c|}{Recorded trajectory}&\multicolumn{3}{c|}{Lateral shift: 3 m}&\multicolumn{3}{c}{Lateral shift: 6 m}\\
\rowcolor{RDTableHeader}
Method&PSNR$\uparrow$&SSIM$\uparrow$&LPIPS$\downarrow$&NTA$\uparrow$&NTL$\uparrow$&FID$\downarrow$&NTA$\uparrow$&NTL$\uparrow$&FID$\downarrow$\\\hline
Street Gaussians~\cite{streetgs}&34.91&0.9528&0.1185&0.2190&48.72&149.50&0.1320&46.57&247.73\\
ReconDreamer~\cite{recondreamer}&33.85&0.9497&0.1235&0.3250&49.72&125.43&0.2460&47.08&197.52\\
ReconDreamer++~\cite{recondreamerpp}&34.69&0.9533&0.1161&0.3650&51.33&118.27&0.2680&49.26&187.51\\\hline
\rowcolor{RDTableHighlight}
\textbf{\method}&\textbf{35.62}&\textbf{0.9592}&\textbf{0.1111}&\textbf{0.3759}&\textbf{51.47}&\textbf{105.41}&\textbf{0.2772}&\textbf{49.53}&\textbf{133.99}\\\RDtablebottomrule
\end{tabular}}
\end{table*}
\begin{figure*}[t]
    \centering
    \includegraphics[width=0.98\textwidth]{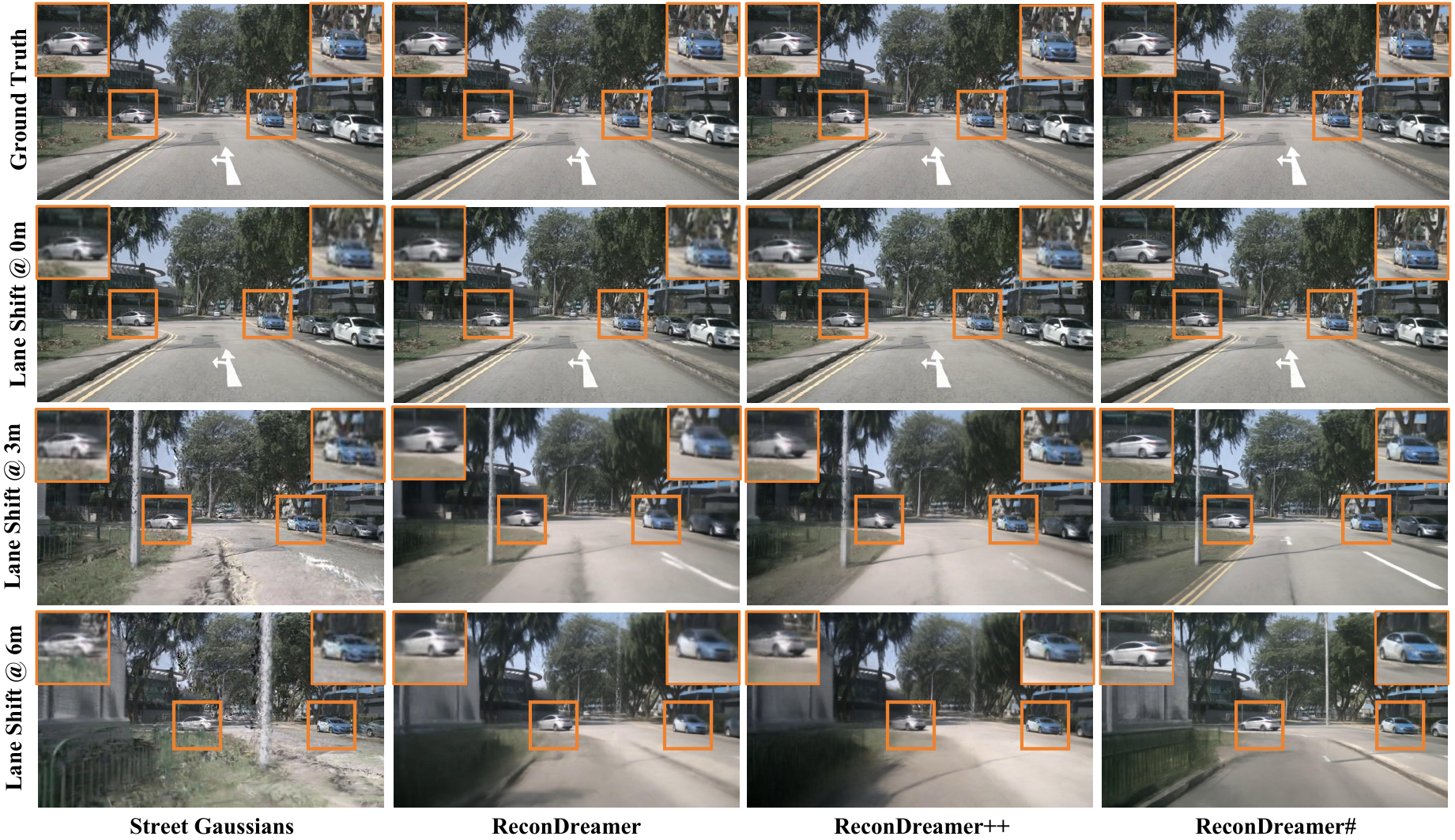}
\caption{Qualitative comparison on nuScenes. Columns show Street Gaussians, ReconDreamer, ReconDreamer++, and \method{}; rows show the recorded-view reference and renderings at 0-m, 3-m, and 6-m lateral shifts. Orange boxes enlarge vehicles and nearby scene details. Ground truth is shown only for the recorded view.}\label{fig:nuscenes}
\end{figure*}
\fi

We organize evaluation around four complementary questions: whether the final scene preserves recorded-view fidelity, whether it remains coherent under trajectory extrapolation, whether AGM and self-evolution contribute independently, and how restoration behavior changes across generations. The following subsections first introduce the experimental setup, then report the main results, and finally present ablations and an efficiency analysis.

\subsection{Experimental Setup}

\textbf{Datasets.} 
We follow ReconDreamer and ReconDreamer++ in dataset selection and conduct experiments on four datasets—Waymo~\cite{waymo}, nuScenes~\cite{nuscenes}, PandaSet~\cite{pandaset,unisim,neurad}, and EUVS~\cite{euvs}—using the same scenes as prior work to ensure fair comparison. Evaluations are conducted on the original trajectory and lateral shifts of 3 meters and 6 meters for Waymo and nuScenes; on the original trajectory, lateral shifts of 2 meters and 3 meters, and a vertical shift of 1 meter for PandaSet; and directly on the training and test sets for EUVS.

\textbf{Metrics.} On Waymo and nuScenes, recorded-view image quality is quantified by PSNR, SSIM~\cite{ssim}, and LPIPS~\cite{lpips}, all of which require aligned reference images. For synthesized trajectories lacking such references, we retain Novel Trajectory Agent IoU (NTA-IoU), Novel Trajectory Lane IoU (NTL-IoU), and FID~\cite{fid} from the prior protocol~\cite{drivedreamer4d,recondreamer}. NTA-IoU measures the consistency of detected agents against projected annotated boxes, whereas NTL-IoU compares lane detections with geometrically transferred lane references. FID compares image distributions and, as such, does not certify the correctness of individual road markings or objects. PandaSet follows prior work~\cite{unisim,neurad} in reporting FID only, while EUVS, given captured images for the extrapolated views, directly reports PSNR, SSIM, and LPIPS.

\textbf{Baselines.} On Waymo, we compare against Street Gaussians~\cite{streetgs}, FreeVS~\cite{freevs}, DriveDreamer4D~\cite{drivedreamer4d}, ReconDreamer~\cite{recondreamer}, and ReconDreamer++~\cite{recondreamerpp}. On nuScenes and EUVS, comparisons are made with Street Gaussians, ReconDreamer, and ReconDreamer++. On PandaSet, we additionally include UniSim~\cite{unisim} and NeuRAD~\cite{neurad}. Notably, FreeVS's LiDAR-conditioned output does not cover the full upper image; we therefore follow the published Waymo crop convention. For each comparison, we adopt the corresponding scene, pose, masking, and metric protocols.

\textbf{Implementation and model selection.} We retain the reconstruction backbone, object decomposition, NTDNet, and sensor-depth supervision. AGM uses a target surface edge length of $0.10$ m, a voxel scale of $0.03$ m, eight neighbors for attribute interpolation, and local LiDAR neighborhoods for plane estimation. These are implementation settings, not constants built into the method or algorithm. The refinement schedule in Algorithm~\floatref{alg:agm} separates the sensor-supported adaptation period from later pseudo-observation updates. For the generation study, we compare selected DriveRestorer generations using the same scene-reconstruction and evaluation protocol.

\subsection{Main Result}
\textbf{Waymo.} Tab.~\floatref{tab:waymo} preserves the complete original- and novel-trajectory comparison. The results show that \method{} improves all nine reported Waymo measures over ReconDreamer++. On recorded views, PSNR increases from 36.29 to 36.89 and LPIPS decreases from 0.1076 to 0.0953, so extrapolation gains do not come at the expense of recorded-view quality. At a three-meter shift, FID falls from 72.02 to 66.42; at six meters, it falls from 111.92 to 94.99. The relative FID reduction grows from 7.8\% to 15.1\% as the shift widens. At six meters, NTA-IoU also rises from 0.489 to 0.532 and NTL-IoU from 56.57 to 58.29, indicating that the distribution-level gain accompanies improved agent and lane consistency. Fig.~\floatref{fig:waymo} shows the corresponding renderings and enlarged vehicle and road details at matched shifts.

\textbf{nuScenes.} Tab.~\floatref{tab:nuscenes} evaluates the same mechanisms under another dataset's cameras and traffic patterns. Relative to ReconDreamer++, \method{} raises recorded-view PSNR from 34.69 to 35.62 and reduces LPIPS from 0.1161 to 0.1111. At a six-meter shift, NTA-IoU increases from 0.268 to 0.2772, NTL-IoU from 49.26 to 49.53, and FID decreases from 187.51 to 133.99. FID also decreases at three meters, from 118.27 to 105.41. The larger six-meter reduction is consistent with the method targeting errors exposed away from the recorded cameras. Fig.~\floatref{fig:nuscenes} shows the shifted-view comparison for vehicles and road appearance.

\textbf{PandaSet.} Tab.~\floatref{tab:pandaset} includes both lateral and vertical pose changes. The vertical test changes the viewing angle over road and object surfaces rather than merely moving the camera across lanes, making it a complementary check of viewpoint extrapolation. \method{} reduces FID relative to ReconDreamer++ from 13.8 to 12.81 without a shift, from 61.9 to 52.73 and 71.7 to 63.8 for two- and three-meter lateral shifts, and from 62.4 to 53.65 for a one-meter vertical shift. Fig.~\floatref{fig:pandaset} illustrates the qualitative improvements under both shift directions.

\ifshowfloats
\begin{table}[!t]\centering\RDtablestyle
\caption{FID comparison on PandaSet under recorded, lateral-shift, and vertical-shift trajectories.}\label{tab:pandaset}
\setlength{\tabcolsep}{3pt}
{\small\begin{tabular}{l|rrrr}\RDtabletoprule
\rowcolor{RDTableHeader}
Method&No shift&Lane 2 m&Lane 3 m&Vert. 1 m\\\hline
UniSim~\cite{unisim}&-&74.70&97.50&-\\
NeuRAD~\cite{neurad}&25.00&72.30&93.90&76.30\\
Street Gaussians~\cite{streetgs}&14.60&66.30&80.70&78.40\\
ReconDreamer~\cite{recondreamer}&16.10&65.40&74.90&67.70\\
ReconDreamer++~\cite{recondreamerpp}&13.80&61.90&71.70&62.40\\\hline
\rowcolor{RDTableHighlight}
\textbf{\method}&\textbf{12.81}&\textbf{52.73}&\textbf{63.80}&\textbf{53.65}\\\RDtablebottomrule
\end{tabular}}
\end{table}
\begin{figure*}[t]\centering
\includegraphics[width=0.98\textwidth,pagebox=cropbox]{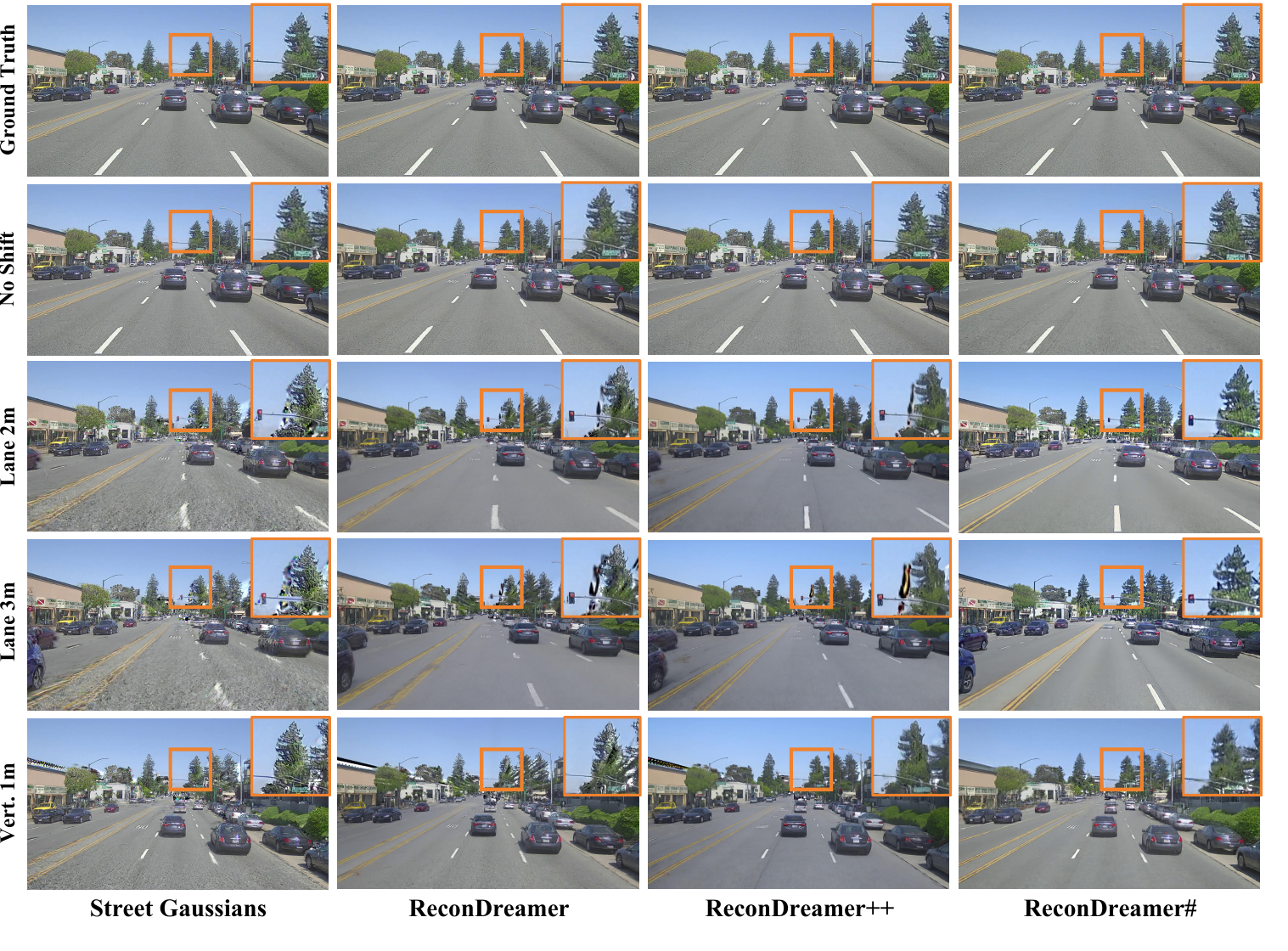}
\caption{Qualitative comparison on PandaSet under recorded, lateral-shift, and vertical-shift trajectories. Columns compare Street Gaussians, ReconDreamer, ReconDreamer++, and \method{}. Orange boxes highlight fine structures in the distant scene.}\label{fig:pandaset}
\end{figure*}
\fi

\textbf{EUVS.} Tab.~\floatref{tab:euvs} measures extrapolated views against real images after foreground masking, complementing the perception-based proxies used for synthesized trajectories. \method{} improves on ReconDreamer++ in all three static-region metrics: PSNR rises from 20.52 to 21.59, SSIM from 0.7033 to 0.724, and LPIPS falls from 0.3057 to 0.2995. Because the test view was independently captured across the lane, these image metrics check alignment to an actual observation rather than only the plausibility of a generated trajectory. Fig.~\floatref{fig:euvs} contrasts the training and test viewpoints; the masking keeps the comparison focused on their shared static content.

\ifshowfloats
\begin{table}[t]\centering\RDtablestyle
\caption{Static-region comparison on EUVS Level 1.}\label{tab:euvs}
\setlength{\tabcolsep}{3.5pt}
{\small\begin{tabular}{l|rrr}\RDtabletoprule
\rowcolor{RDTableHeader}
Method&PSNR$\uparrow$&SSIM$\uparrow$&LPIPS$\downarrow$\\\hline
Street Gaussians~\cite{streetgs}&15.08&0.6043&0.3428\\
ReconDreamer~\cite{recondreamer}&20.14&0.6914&0.3286\\
ReconDreamer++~\cite{recondreamerpp}&20.52&0.7033&0.3057\\\hline
\rowcolor{RDTableHighlight}
\textbf{\method}&\textbf{21.59}&\textbf{0.7240}&\textbf{0.2995}\\\RDtablebottomrule
\end{tabular}}
\end{table}
\begin{figure*}[t]\centering
\includegraphics[width=0.98\textwidth,pagebox=cropbox]{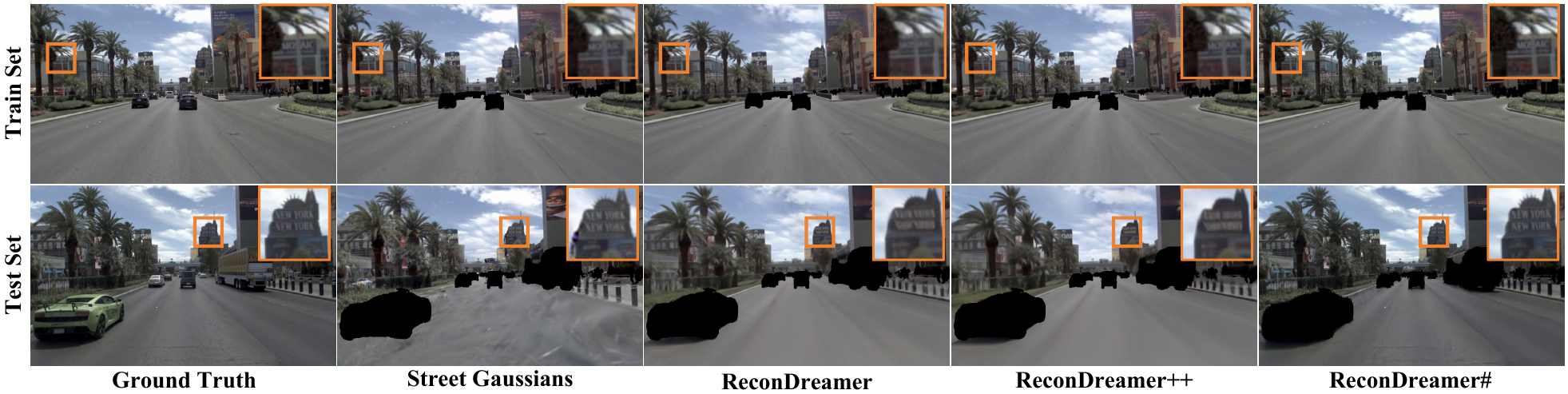}
\caption{Qualitative comparison on EUVS Level 1. The upper row shows a training-trajectory view (Train Set), while the lower row shows an independently captured cross-lane test view (Test Set). Columns compare ground truth, Street Gaussians, ReconDreamer, ReconDreamer++, and \method{}. Dynamic foreground is masked in the rendered comparisons, and orange boxes enlarge facade signs.}\label{fig:euvs}
\end{figure*}
\fi

\subsection{Module Ablation Studies}
We isolate the two extensions introduced by \method{}, including the AGM and the self-evolving reconstruction--restoration framework.

\ifshowfloats
\begin{table*}[t]\centering\RDtablestyle
\caption{New-module ablation on Waymo. SE denotes Self-Evolving DriveRestorer. All variants retain the inherited depth and NTDNet components.}\label{tab:modules}
{\small\begin{tabular}{l|cc|rrr|rrr|rrr}\RDtabletoprule
\rowcolor{RDTableHeader}
&SE&AGM&\multicolumn{3}{c|}{Recorded}&\multicolumn{3}{c|}{3 m shift}&\multicolumn{3}{c}{6 m shift}\\
\rowcolor{RDTableHeader}
Method&&&PSNR$\uparrow$&SSIM$\uparrow$&LPIPS$\downarrow$&NTA$\uparrow$&NTL$\uparrow$&FID$\downarrow$&NTA$\uparrow$&NTL$\uparrow$&FID$\downarrow$\\\hline
ReconDreamer++&&&36.29&0.9573&0.1076&0.5720&57.06&72.02&0.4890&56.57&111.92\\
+ AGM&&\checkmark&36.72&0.9608&0.1015&0.5780&58.00&71.72&0.5070&57.40&108.10\\
+ Self-Evolution&\checkmark&&36.43&0.9568&0.1072&0.5830&57.75&70.45&0.5120&55.95&100.11\\
\hline
\rowcolor{RDTableHighlight}
\textbf{\method}&\checkmark&\checkmark&\textbf{36.89}&\textbf{0.9638}&\textbf{0.0953}&\textbf{0.5984}&\textbf{58.75}&\textbf{66.42}&\textbf{0.5320}&\textbf{58.29}&\textbf{94.99}\\\RDtablebottomrule
\end{tabular}}
\end{table*}
\begin{figure*}[t]\centering
\includegraphics[width=0.98\textwidth,pagebox=cropbox]{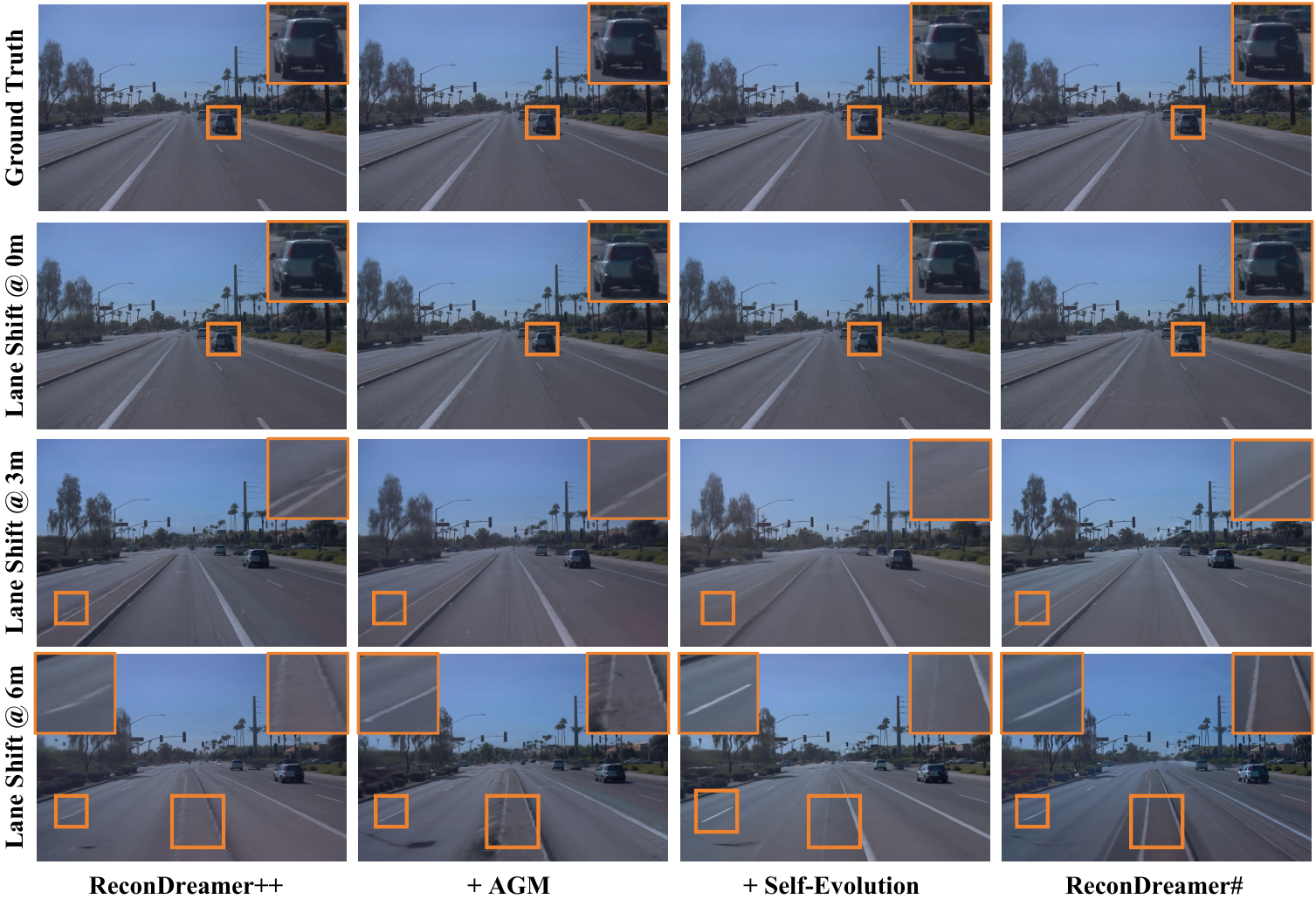}
\caption{Qualitative module ablation on Waymo. Columns compare ReconDreamer++, +AGM, +Self-Evolution, and the complete \method{} model. Rows show the reference view and renderings at 0~m, 3~m, and 6~m lateral shifts. Orange boxes highlight vehicle details and road-boundary artifacts.}\label{fig:ablation}
\end{figure*}
\fi

Tab.~\floatref{tab:modules} implements a two-factor comparison. The AGM-only variant uses the same fixed restoration model as the base, and the self-evolution-only variant retains the original ground treatment. AGM alone improves recorded-view LPIPS from 0.1076 to 0.1015 and six-meter NTL-IoU from 56.57 to 57.4. Self-evolution alone reduces six-meter FID from 111.92 to 100.11, although its six-meter NTL-IoU falls to 55.95. This contrast matters: better generated appearance does not by itself preserve lane structure at the larger shift. Combining both yields the best value in every reported column, including 0.0953 LPIPS, 58.29 six-meter NTL-IoU, and 94.99 six-meter FID.

For a higher-is-better metric $q$, complementarity can be summarized by the difference of differences,
\begin{equation}
\Delta_{\rm int}=q_{\rm both}-q_{\rm SE}-q_{\rm AGM}+q_{\rm base}.
\end{equation}
This is an analysis statistic, not an optimization objective or a guarantee of synergy. For lower-is-better metrics, the sign convention is reversed before interpretation. On six-meter NTL-IoU, $\Delta_{\rm int}=1.51$, indicating that the combined gain exceeds the sum of the isolated gains for this measure. Fig.~\floatref{fig:ablation} shows the four variants at matched poses, with enlarged vehicle and road-boundary regions.

\subsection{Multi-Round DriveRestorer Analysis}
In this section, we analyze in detail the substantial gains brought by the self-evolving framework to DriveRestorer. As shown in Tab.~\floatref{tab:generations}, $\mathcal R_1$ is the initial static restorer and $\mathcal R_2$--$\mathcal R_4$ correspond to successive self-evolution generations, consistent with the notation in Sec.~\ref{sec:method}. To ensure a fair and directly comparable assessment, each restorer generation is evaluated by integrating it into the complete reconstruction pipeline and measuring the quality of the final reconstructed scene, so that the reported results reflect the effect of the restorer itself rather than that of any auxiliary component.

\ifshowfloats
\begin{table*}[t]\centering\RDtablestyle
\caption{Multi-Round DriveRestorer comparison on Waymo. $\mathcal R_1$ is the initial static restorer; $\mathcal R_2$--$\mathcal R_4$ are successive self-evolution generations.}\label{tab:generations}
{\small\begin{tabular}{l|r|rrr|rrr}\RDtabletoprule
\rowcolor{RDTableHeader}
 &Recorded &\multicolumn{3}{c|}{3 m shift}&\multicolumn{3}{c}{6 m shift}\\
\rowcolor{RDTableHeader}
DriveRestorer&LPIPS$\downarrow$&NTA$\uparrow$&NTL$\uparrow$&FID$\downarrow$&NTA$\uparrow$&NTL$\uparrow$&FID$\downarrow$\\\hline
$\mathcal R_1$ &0.1015&0.5780&58.00&71.72&0.5070&57.40&108.10\\
$\mathcal R_2$ &0.0985&0.5880&58.34&68.96&0.5180&57.84&101.94\\
$\mathcal R_3$ &0.0974&0.5910&58.49&68.17&0.5240&57.99&99.45\\
\hline
\rowcolor{RDTableHighlight}
\textbf{$\mathcal R_4$ }&\textbf{0.0953}&\textbf{0.5984}&\textbf{58.75}&\textbf{66.42}&\textbf{0.5320}&\textbf{58.29}&\textbf{94.99}\\\RDtablebottomrule
\end{tabular}}
\end{table*}
\fi

With AGM held fixed, each successive restorer generation improves all reported aggregate measures. From $\mathcal R_1$ to $\mathcal R_4$, recorded-view LPIPS decreases from 0.1015 to 0.0953; at a six-meter shift, NTA-IoU increases from 0.507 to 0.532, NTL-IoU from 57.4 to 58.29, and FID decreases from 108.10 to 94.99. The three-meter FID likewise drops from 71.72 to 66.42, indicating that the later restorers improve both tested shift distances under the fixed ground policy. These aggregate trends support reconstruction-mediated pair mining, while not implying that every scene or future generation must improve monotonically.

\subsection{Ground Representation Analysis}
We further examine how AGM changes the spatial distribution and local surface alignment of the ground representation.

\ifshowfloats
\begin{figure*}[!t]\centering
\includegraphics[width=0.99\textwidth]{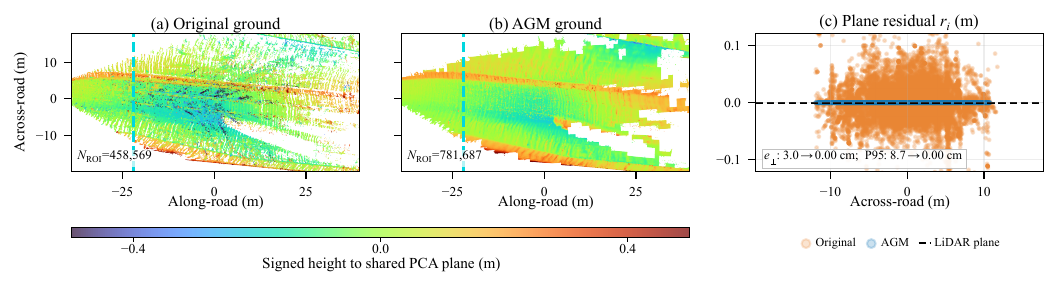}
\caption{Ground-support comparison on Waymo scene 065. (a) The original ReconDreamer++ ground uses fixed support. (b) Complete AGM produces denser and more uniform support. Both top views share the same bounds, point size, and height colors; the cyan line marks the cross-section. (c) Signed residual $r_i=n_i^\top(\mu_i-p_i)$ to the local LiDAR plane. The mean absolute residual $e_\perp$ and P95 show that AGM expands road support while maintaining surface alignment.}\label{fig:ground}
\end{figure*}
\fi

Fig.~\floatref{fig:ground} complements rendered images with a direct view of ground support. Within the plotted region, the ground-primitive count rises from 458,569 to 781,687, while the shared top-view layout shows fewer holes and more even road coverage. More primitives alone would not establish a better ground surface: the matched local profile also checks whether their centers remain near the LiDAR plane. We quantify this property by the sensor-plane residual $e_\perp=N_g^{-1}\sum_i|n_i^\top(\mu_i-p_i)|$. The concentrated residual profile alongside the increased coverage shows that AGM expands support without a corresponding spread away from the sensor-aligned road surface.

\subsection{Efficiency, Limitations, and Discussion}
\raggedbottom
Self-evolution introduces offline computation for novel-video restoration, auxiliary reconstruction, round-trip rendering, and restorer updating. AGM adds surface reconstruction and neighborhood estimation, and support preservation can increase the final primitive count. These training costs are distinct from the rendering throughput of the optimized Gaussian scene. The auxiliary reconstruction is not queried during ordinary final-scene rasterization, while any online restoration still incurs the cost of the generative model.

The method inherits several limitations. Round-trip mining can only supervise consequences visible from the recorded trajectory, and it cannot independently verify unseen geometry. Local ground planes can be unreliable near abrupt elevation changes or erroneous segmentation. Preserving low-opacity support may retain redundant primitives, and a surface-based initialization can bridge regions that should remain disconnected if the input geometry is wrong. Generative restoration may produce a coherent but incorrect appearance, especially for details not supported by the observations. These limitations motivate retaining real-image evaluation and sensor-derived geometry throughout the comparison.

The combined design provides a division of responsibility rather than a universal correction guarantee. The restorer adapts to encountered degradation, the auxiliary model makes some of that degradation supervisable, and AGM constrains how road support changes. The controlled ablations show that neither extension alone matches the complete model across recorded and shifted views, while the generation study shows consistent aggregate gains through $\mathcal R_4$. The remaining failure modes concern regions not observed by the round-trip path and geometry not reliably constrained by LiDAR.

\section{Conclusion}\label{sec:conclusion}
We presented \method{}, which extends generative--reconstructive driving-scene modeling through Self-Evolving DriveRestorer and an Adaptive Ground Model. The self-evolution procedure restores novel-trajectory renders, fits an auxiliary scene, and returns to recorded poses to construct restoration inputs with authentic image targets. This changes the degradation distribution available to successive restoration generations while preserving a supervised target domain. AGM complements that process by constructing surface-based road support, restricting split displacements to local tangential directions, projecting support onto LiDAR planes, and removing opacity-only ground deletion. Its formulation distinguishes adaptive support changes from continuous geometry and appearance optimization.

The resulting framework makes restoration learning responsive to the reconstruction it supports and keeps ground adaptation tied to sensor evidence. Its evaluation spans original trajectories, synthesized lane shifts, vertical perturbations, and independently captured cross-lane views. Relative to ReconDreamer++, the complete model reduces Waymo FID from 111.92 to 94.99 at a 6-m shift while increasing NTA-IoU from 0.489 to 0.532; it also reduces nuScenes FID from 187.51 to 133.99 at the same shift. The module and generation ablations show complementary gains from AGM and successive restorer updates. Remaining challenges include failures invisible to the round-trip path, inaccurate local ground estimates, and balancing support preservation against computational cost.

\ifdefined\NOBIB
\else
\bibliographystyle{IEEEtran}
\bibliography{references}
\fi

\end{document}